\documentclass[conference]{IEEEtran}
\usepackage{cite}

\ifCLASSINFOpdf
  \usepackage[pdftex]{graphicx}
\else
\fi
\usepackage{multirow}
\usepackage[table,xcdraw]{xcolor}

\begin{document}
%
\title{Analysing object detectors from the perspective of co-occurring object categories}

\author{\IEEEauthorblockN{Csaba~Nemes}
\IEEEauthorblockA{Nokia Bell Labs\\
Budapest, Hungary\\
Email: csaba.nemes@nokia-bell-labs.com}
\and
\IEEEauthorblockN{Sandor~Jordan}
\IEEEauthorblockA{Nokia Bell Labs\\
Budapest, Hungary\\
Email: sandor.jordan@nokia-bell-labs.com}
}


%


\maketitle

\begin{abstract}
The accuracy of state-of-the-art Faster R-CNN and YOLO object detectors are evaluated and compared on a special masked MS COCO dataset to measure how much their predictions rely on contextual information encoded at object category level. Category level representation of context is motivated by the fact that it could be an adequate way to transfer knowledge between visual and non-visual domains.
According to our measurements, current detectors usually do not build strong dependency on contextual information at category level, however, when they does, they does it in a similar way, suggesting that contextual dependence of object categories is an independent property that is relevant to be transferred.

\end{abstract}

\begin{IEEEkeywords}
Deep learning, co-occurance, object detection, image recognition, MS COCO, R-CNN, YOLO\end{IEEEkeywords}

%
\IEEEpeerreviewmaketitle

\section{Introduction}

One of the primary goals of Cognitive Informatics is to boost the efficiency of human-machine interactions~\cite{trends}, for which it is inevitable to develop efficient knowledge representations that can be utilized in multiple domains.
For example, a machine agent shall be able to learn from human instructions (NLP domain) and recognize surrounding objects (visual domain) as well.
Hence, knowledge engineering~\cite{knowledgeEngineering}, concept formulation~\cite{conceptFormation}, or knowledge transfer~\cite{transferLEarning} can be regarded as elementary building blocks of the field.

In the paper, object detection is chosen as an example visual task to investigate what knowledge can be gained from a visual dataset, that can be re-used in a non-visual task. As a candidate, co-occurrence statistic of objects is investigated given our intuition that typically the same objects co-occur in images that co-occur in texts (describing scenes). For example, \textit{chair} and \textit{table} frequently co-occur in both images and in sentences.

The goal of neural network based object detection~\cite{renNIPS15fasterrcnn,Redmon_2017_CVPR} is to approximate the image patch \& label statistic encoded in the training set in such a way that generalizes well in the rest of the domain.

To avoid training of large networks from scratch, networks are usually initialized from another network which has already been trained for the same domain or another highly related domain. The technique is called transfer learning~\cite{pan2010survey} as its purpose is to carry over statistics from one domain to another.

Similar domains can cover only one portion of the knowledge we can acquire about the world, hence to surpass current object detection techniques the transfer between non-related domains should be addressed.
More specifically, one should ask what statistics can be extracted from non-visual domains which can be used in visual tasks and what statistics can be extracted from visual domains which can be used in a non-visual tasks.

In case of object detection, one transferable statistic can be the statistic of object categories co-occurring in the same image. 
Co-occurring object categories can represent the environment of an object which can be used to fine-tune object detection, as in numerous cases the classification solely on object pixels is ambivalent.
Assuming images are the projections of complex scenes which are also referenced in texts and knowledge bases, co-occurring category statistic can be extracted from non-visual domains as well.

In the paper, the performance of two state-of-the-art object detectors~\cite{renNIPS15fasterrcnn,Redmon_2017_CVPR} is evaluated on the MS COCO dataset~\cite{mscoco14} to investigate how much the detectors rely on the object pixels, and how much they deduct from the pixels of co-occurring object categories.

We can assume current detectors not only learn the pixel patterns of an object category but at some extent the context of category as well.
\begin{itemize}
    \item If they are not using the environmental information described at category level, a transfer technique may improve their performance in complex cases where the accuracy is not good enough.
    \item If they are already using such environmental information, the learned information should be compared to ones acquired from different domains. Harmonizing statistics could improve generalization in all domains.
\end{itemize}

For each object category, a masked dataset is created in which the instances of the category are masked out with grey color. On the masked datasets, the detection performance is recorded for each category. Comparison to the unmasked performance reveals how much the detectors use the presence of a surrounding object category. Highest impact contextual categories reported for each category. If similar category pairs are found for both detectors, that suggests these properties belongs to the dataset and not to the detectors themselves.

In Section~\ref{related}, related works motivating the co-occurrence statistic is described.
In Section~\ref{dataset}, the MS COCO dataset and our masking process are presented.
In Section~\ref{implementation}, the object detectors used in the evaluation are summarized.
Finally the results and the conclusion are given in Section~\ref{results} and~\ref{conclusion}, respectively.

\begin{figure}[!b]
\centering
\includegraphics[width=0.8\linewidth]{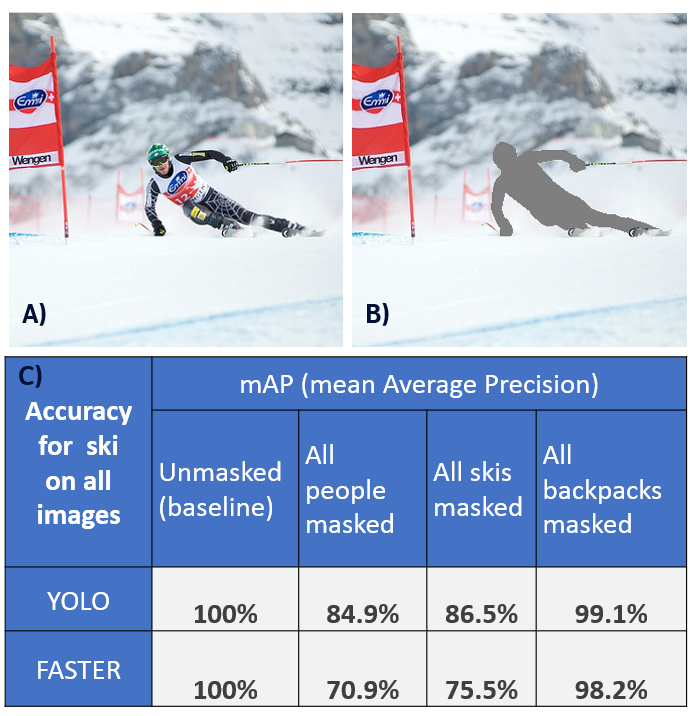}
\caption{\textbf{A)} Unmasked image. \textbf{B)} Image with people masked out. \textbf{C)} Accuracy of both detectors drastically decreases if people around skis are masked out. This suggests that skis are ambivalent pixel patches, and detectors need the presence of a person in the image to fine-tune the prediction.
}
\label{fig_skis}
\end{figure}

\section{Related works}\label{related}

Looking at co-occurring category statistic was motivated by the success of hierarchically structured labels~\cite{StructuredLabel}. They observed that some labels (aka categories) occur together more frequently than others. They proposed to use scene types to represent the most common co-occurrences, hence it can be interpreted as a special case of our proposal. Their method contained separate networks to detect the most relevant scene types and than for each scene type a separate network to detect the relevant objects. Their method can also be interpreted as a transfer technique in which the human insight is transferred into to detector in the form of the user defined scenes and their relevant object categories.

Our approach is different, because we consider object-object co-occurrence explicitly while they build on object-scene statistics.

In \cite{TransferCommmonSenseKnowledge}, a complex technique is presented for a special case of transfer learning, in which knowledge learned on source categories (with bounding boxes) is transferred to similar target categories, where only image level categories are available. It combines the already learned source categories with common-sense knowledge automatically acquired from knowledge bases to learn new categories.
Two of their knowledge bases can be related to our measurements: (i) scene and (ii) spatial common-sense.

Scene common-sense approach is a re-implementation of the hierarchically structured labels approach, hence, it takes the scenes from an external resource and does not consider the object-object statistics directly.

Spatial common-sense, however, goes one step further.
It acquires (category1, category2, spatial relation) statistics from a knowledge base with relational annotations.
Beside object categories co-occurring in the same image it considers their spatial relation as well to improve object detection of target classes.


Related works, in general, focus on how to improve state-of-the-art object detectors with co-occurrence statistics acquired from 3rd party knowledge bases, while it has not been investigated how much these detectors rely on co-occurrence statistics encoded in the dataset itself.
Our paper aims to fill the gap by analyzing how much these detectors  use from the co-occurrence statistics encoded in the MS-COCO dataset.
If relevant statistics can be extracted from the dataset it opens the door to reverse the direction of the information flow by improving knowledge bases or by reusing these statistics in non-visual tasks.



\section{Dataset}\label{dataset}

\subsection{MS COCO}

Our evaluation was done on The Microsoft COCO 2014 dataset~\cite{COCO}, which compared to ImageNet~\cite{imagenet} contains more complex scenes with multiple objects in it. It contains more than 150K images divided into train, validation and test sets. During training all the training data and 35K images selected from the validation set were used. For validation and masking, we used the minival2014 dataset~\cite{minival2014} which contains the remaining 5K images of the validation set. This technique was proposed to enable a quick evaluation~\cite{minvalPaper} which approximates measurements on the test set relatively well.


\subsection{Masking process}

The core idea of our analysis is to present specially masked images to the detectors to test how their accuracy decreases when all instances of a given category is masked out.

We created a masked dataset for all 80 object categories, by finding the segmentation mask of each instance of the given category in the annotation file and setting the color of the pixels of the segmentation mask to grey.

During the masking process, if an area to be masked out had been overlapping with an object of a different category we carefully skipped the overlapping area from the mask. Hence, the deleted amount of pixels of the chosen category was equal or smaller, than the number of pixels corresponding to that object category. 

\begin{figure}[!b]
\centering
\includegraphics[width=0.8\linewidth]{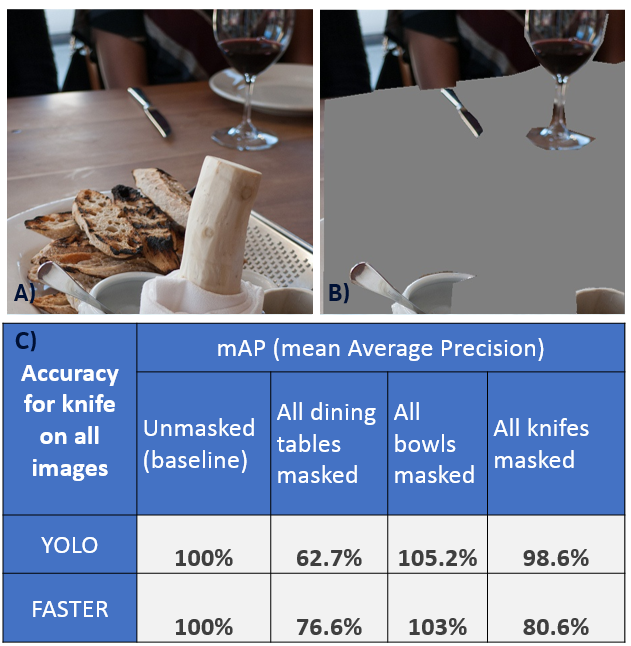}
\caption{\textbf{A)} Unmasked image. \textbf{B)} Image with dining tables masked out. \textbf{C)} Accuracy of both detectors drastically decreases if dining tables around knives are masked out, however, it increases if bowls are removed from the images.}
\label{fig_knife}
\end{figure}

Examples for masks are shown in Figure~\ref{fig_skis} and~\ref{fig_knife}.

\section{Implementation details}\label{implementation}

\subsection{Faster R-CNN}

Region-based Convolutional Neural Networks (R-CNN) belong to the two-stage object detectors. These detectors depends on a region proposal subsystem to identify the location of the object on the image, and an object recognition subsystem to classify the detected object. We used a state-of-the-art implementation~\cite{wu2016tensorpack} of Faster R-CNN~\cite{renNIPS15fasterrcnn}. The Faster-RCNN does not contain any external region proposal network (RPN) as in the case of R-CNN or Fast R-CNN, but it uses its own CNN layers to propose regions from a convolutional feature map. Based on the proposed regions and the feature map it can determine the object category.

A Faster R-CNN implementation based on a ResNet-50 model~\cite{HeZRS15} was trained on COCO trainval35k and reached a mean Average Precision (mAP) of $0.369$ at IoU=0.50:0.95. 

\subsection{YOLO}
Unlike Faster R-CNN, YOLO is a one-stage object detector ~\cite{YOLO1}. It means YOLO architecture consists of a single neural network, which determines the bounding boxes of the objects and associates the class label for each bounding box in one evaluation cycle. The accuracy of the algorithm is usually lower than in the case of Faster R-CNNs, however, YOLO is mush faster. Compared to other real-time detectors, like DPM ~\cite{DPM}, YOLO has the best accuracy.

For our experiments we used the Darkflow ~\cite{DarkFlow} implementation of YOLO. The applied architecture follows YOLO V2 ~\cite{DarkNet}. The model was trained on COCO trainval35k ~\cite{YOLO9000}. Input images were re-sized to $608$x$608$. We evaluated the overall accuracy of the applied model on the COCO minival2014 dataset. The average precision of the applied YOLO V2 network using IoU=0.50:0.95 metric was $0.225$. The evaluation was carried out with the COCO API ~\cite{COCOAPI}. Faster R-CNN can generate higher accuracy than YOLO V2, but it is an offline algorithm, since it takes ~2 sec to process one image while YOLO V2 can work at $\sim 40$ FPS.

\section{Results}\label{results}

\subsection{AP (Average Precision) for each object category}

\begin{figure}[!b]
\centering
\includegraphics[width=0.9\linewidth]{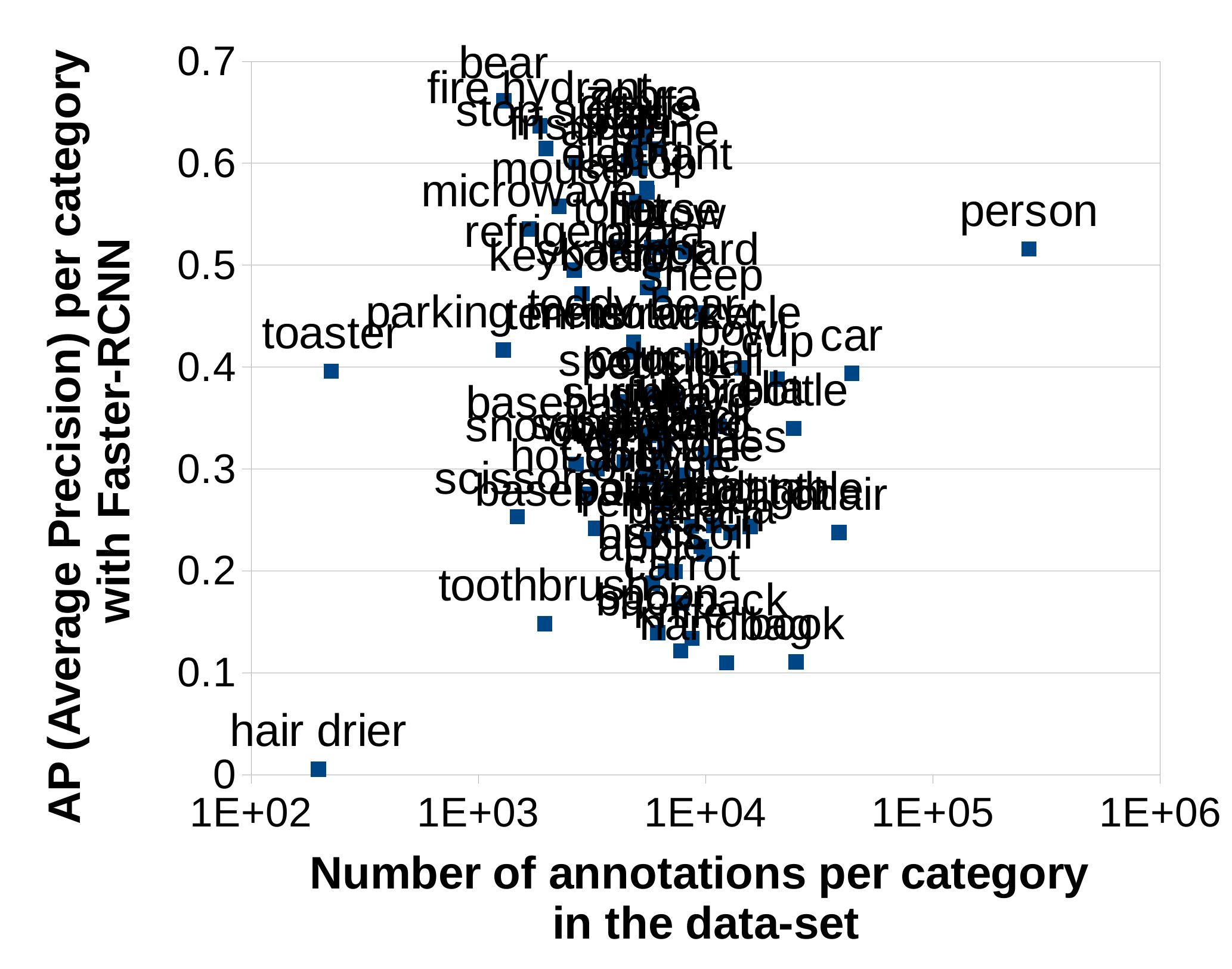}
\caption{Faster-RCNN: AP per each category plotted versus the number of annotations per category.
}
\label{fig_general}
\end{figure}

Average Precision per each category is measured and plotted versus the number of annotations (instances) per object category in Figure~\ref{fig_general}. In general, the accuracy of the detectors varies with the categories, however the variance does not correlate with the number of available annotations. This suggests that the dataset contains object categories with various complexity. (For complex categories even high number of annotations are not sufficient.)

\subsection{Masking results for the top-10 object categories based on detection accuracy}

For the top-10 object categories based on detection accuracy, the accuracy drops of Faster-RCNN and YOLO on the masked datasets are displayed in
Table~\ref{label_table_faster_top10} and Table~\ref{label_table_yolo_top10}, respectively.

For each category, only measurements on those masked datasets are reported which belong to the 3 largest deviations in accuracy. These measurements are labeled by the category which was masked out in the given measurement. The displayed categories indicate which surrounding objects affect the detector performance the most.

For these categories, the detector performs quite well, and does not rely much on contextual information represented at category level (except frisbee at Faster-RCNN). Apparently, the pixel patches of these categories can be easily detected without contextual information.

\begin{table}
\centering
\caption{For the top-10 object categories based on detection accuracy, the accuracy drop of Faster-RCNN is reported. For each category, only those masked datasets are reported which belongs to the 3 largest deviations in accuracy. Datasets are labeled by the category which was masked out.}
\label{label_table_faster_top10}
\begin{tabular}{|
>{\columncolor[HTML]{C0C0C0}}l |l|c|c|c|}
\hline
\cellcolor[HTML]{EFEFEF}{\color[HTML]{333333} } & \cellcolor[HTML]{EFEFEF}{\color[HTML]{333333} } & \multicolumn{3}{c|}{\cellcolor[HTML]{EFEFEF}{\color[HTML]{333333} \textbf{\begin{tabular}[c]{@{}c@{}}Top-3 masked data sets for each \\ category which altered AP \\ the most (AP change in \%)\end{tabular}}}} \\ \cline{3-5} 
\multirow{-3}{*}{\cellcolor[HTML]{EFEFEF}{\color[HTML]{333333} \textbf{\begin{tabular}[c]{@{}l@{}}Categories\\ with top \\ accuracy\\ (AP)\end{tabular}}}} & \multirow{-3}{*}{\cellcolor[HTML]{EFEFEF}{\color[HTML]{333333} \textbf{\begin{tabular}[c]{@{}l@{}}AP\\ for each\\ category\end{tabular}}}} & \cellcolor[HTML]{EFEFEF}{\color[HTML]{333333} 1st} & \cellcolor[HTML]{EFEFEF}{\color[HTML]{333333} 2nd} & \cellcolor[HTML]{EFEFEF}{\color[HTML]{333333} 3rd} \\ \hline
bear & 0,661470737 & \begin{tabular}[c]{@{}c@{}}bear \\ (0.1\%)\end{tabular} & \begin{tabular}[c]{@{}c@{}}dog \\ (101.1\%)\end{tabular} & \begin{tabular}[c]{@{}c@{}}cat \\ (100.5\%)\end{tabular} \\ \hline
fire hydrant & 0,636703187 & \begin{tabular}[c]{@{}c@{}}fire hydrant\\ (8.8\%)\end{tabular} & \begin{tabular}[c]{@{}c@{}}person \\ (97.6\%)\end{tabular} & \begin{tabular}[c]{@{}c@{}}car \\ (100.8\%)\end{tabular} \\ \hline
zebra & 0,629035111 & \begin{tabular}[c]{@{}c@{}}zebra\\ (0.0\%)\end{tabular} & \begin{tabular}[c]{@{}c@{}}car\\ (100.2\%)\end{tabular} & \begin{tabular}[c]{@{}c@{}}truck \\ (100.2\%)\end{tabular} \\ \hline
giraffe & 0,620582563 & \begin{tabular}[c]{@{}c@{}}giraffe\\ (18.6\%)\end{tabular} & \begin{tabular}[c]{@{}c@{}}zebra\\ (101.1\%)\end{tabular} & \begin{tabular}[c]{@{}c@{}}car \\ (100.9\%)\end{tabular} \\ \hline
bus & 0,61581961 & \begin{tabular}[c]{@{}c@{}}bus\\ (6.5\%)\end{tabular} & \begin{tabular}[c]{@{}c@{}}car\\ (97.3\%)\end{tabular} & \begin{tabular}[c]{@{}c@{}}truck \\ (101.4\%)\end{tabular} \\ \hline
stop sign & 0,614616556 & \begin{tabular}[c]{@{}c@{}}stop sign \\ (57.5\%)\end{tabular} & \begin{tabular}[c]{@{}c@{}}person\\ (98.9\%)\end{tabular} & \begin{tabular}[c]{@{}c@{}}truck \\ (99.3\%)\end{tabular} \\ \hline
cat & 0,61117558 & \begin{tabular}[c]{@{}c@{}}cat\\ (11.1\%)\end{tabular} & \begin{tabular}[c]{@{}c@{}}bed \\ (101.1\%)\end{tabular} & \begin{tabular}[c]{@{}c@{}}person \\ (99.0\%)\end{tabular} \\ \hline
train & 0,603173341 & \begin{tabular}[c]{@{}c@{}}train\\ (1.5\%)\end{tabular} & \begin{tabular}[c]{@{}c@{}}handbag \\ (99.2\%)\end{tabular} & \begin{tabular}[c]{@{}c@{}}bus \\ (100.8\%)\end{tabular} \\ \hline
frisbee & 0,600818055 & \begin{tabular}[c]{@{}c@{}}frisbee\\ (40.8\%)\end{tabular} & \cellcolor[HTML]{FFFFC7}\begin{tabular}[c]{@{}c@{}}person\\ (84.4\%)\end{tabular} & \begin{tabular}[c]{@{}c@{}}dog \\ (96.1\%)\end{tabular} \\ \hline
airplane & 0,595326524 & \begin{tabular}[c]{@{}c@{}}airplane\\ (44.3\%)\end{tabular} & \begin{tabular}[c]{@{}c@{}}person \\ (97.6\%)\end{tabular} & \begin{tabular}[c]{@{}c@{}}truck \\ (98.9\%)\end{tabular} \\ \hline
\end{tabular}
\end{table}

\begin{table}
\centering
\caption{Similar to Table~\ref{label_table_faster_top10} in case of YOLO.} 
\label{label_table_yolo_top10}
\begin{tabular}{|
>{\columncolor[HTML]{C0C0C0}}l |l|c|c|c|}
\hline
\cellcolor[HTML]{EFEFEF}{\color[HTML]{333333} } & \cellcolor[HTML]{EFEFEF}{\color[HTML]{333333} } & \multicolumn{3}{c|}{\cellcolor[HTML]{EFEFEF}{\color[HTML]{333333} \textbf{\begin{tabular}[c]{@{}c@{}}Top-3 masked data sets for each \\ category which altered AP \\ the most (AP change in \%)\end{tabular}}}} \\ \cline{3-5} 
\multirow{-3}{*}{\cellcolor[HTML]{EFEFEF}{\color[HTML]{333333} \textbf{\begin{tabular}[c]{@{}l@{}}Categories\\ with top \\ accuracy\\ (AP)\end{tabular}}}} & \multirow{-3}{*}{\cellcolor[HTML]{EFEFEF}{\color[HTML]{333333} \textbf{\begin{tabular}[c]{@{}l@{}}AP\\ for each\\ category\end{tabular}}}} & \cellcolor[HTML]{EFEFEF}{\color[HTML]{333333} 1st} & \cellcolor[HTML]{EFEFEF}{\color[HTML]{333333} 2nd} & \cellcolor[HTML]{EFEFEF}{\color[HTML]{333333} 3rd} \\ \hline
\textbf{bear} & 0,576679984 & \begin{tabular}[c]{@{}c@{}}bear \\ (0.0\%)\end{tabular} & \begin{tabular}[c]{@{}c@{}}bird \\ (100.8\%)\end{tabular} & \begin{tabular}[c]{@{}c@{}}cow\\ (100.3\%)\end{tabular} \\ \hline
\textbf{giraffe} & 0,478006143 & \begin{tabular}[c]{@{}c@{}}giraffe\\ (20.6\%)\end{tabular} & \begin{tabular}[c]{@{}c@{}}zebra\\ (100.6\%)\end{tabular} & \begin{tabular}[c]{@{}c@{}}person\\ (100.2\%)\end{tabular} \\ \hline
train & 0,463421221 & \begin{tabular}[c]{@{}c@{}}train\\ (1.6\%)\end{tabular} & \begin{tabular}[c]{@{}c@{}}bus\\ (101.2\%)\end{tabular} & \begin{tabular}[c]{@{}c@{}}person \\ (101.0\%)\end{tabular} \\ \hline
stop sign & 0,463144531 & \begin{tabular}[c]{@{}c@{}}stop sign \\ (33.9\%)\end{tabular} & \begin{tabular}[c]{@{}c@{}}car \\ (98.1\%)\end{tabular} & \begin{tabular}[c]{@{}c@{}}cow\\ (98.1\%)\end{tabular} \\ \hline
toilet & 0,461524673 & \begin{tabular}[c]{@{}c@{}}toilet \\ (41.9\%)\end{tabular} & \begin{tabular}[c]{@{}c@{}}cat \\ (98.1\%)\end{tabular} & \begin{tabular}[c]{@{}c@{}}dog\\ (98.9\%)\end{tabular} \\ \hline
elephant & 0,460009161 & \begin{tabular}[c]{@{}c@{}}elephant \\ (8.8\%)\end{tabular} & \begin{tabular}[c]{@{}c@{}}person \\ (98.8\%)\end{tabular} & \begin{tabular}[c]{@{}c@{}}bench\\ (99.5\%)\end{tabular} \\ \hline
zebra & 0,458991063 & \begin{tabular}[c]{@{}c@{}}zebra\\ (0.0\%)\end{tabular} & \begin{tabular}[c]{@{}c@{}}giraffe \\ (99.8\%)\end{tabular} & \begin{tabular}[c]{@{}c@{}}car\\ (100.2\%)\end{tabular} \\ \hline
bus & 0,455744453 & \begin{tabular}[c]{@{}c@{}}bus \\ (4.8\%)\end{tabular} & \begin{tabular}[c]{@{}c@{}}car \\ (96.6\%)\end{tabular} & \begin{tabular}[c]{@{}c@{}}person\\ (97.6\%)\end{tabular} \\ \hline
airplane & 0,452356588 & \begin{tabular}[c]{@{}c@{}}airplane\\ (57.1\%)\end{tabular} & \begin{tabular}[c]{@{}c@{}}truck\\ (99.2\%)\end{tabular} & \begin{tabular}[c]{@{}c@{}}giraffe \\ (99.3\%)\end{tabular} \\ \hline
cat & 0,437964013 & \begin{tabular}[c]{@{}c@{}}cat\\ (9.2\%)\end{tabular} & \begin{tabular}[c]{@{}c@{}}bed \\ (102.6\%)\end{tabular} & \begin{tabular}[c]{@{}c@{}}sink\\ (101.9\%)\end{tabular} \\ \hline

\end{tabular}
\end{table}

\subsection{Masking results for the categories which have the largest contextual dependence}

For categories with the largest contextual dependence, the accuracy drops of Faster-RCNN and YOLO are displayed in
Table~\ref{label_table_faster_affected} and Table~\ref{label_table_yolo_affected}, respectively.

For these categories, the accuracy is heavily affected if certain object categories are masked out in the images. Moreover, in the highlighted cases (yellow cells) masking the context category affects the accuracy more than masking of the object itself. This suggest that these image patches are so ambivalent in the dataset that more information is encoded in their environment. In this cases the detectors learned how to use the contextual statistics at the level of categories. These are the candidates to be compared to statistics acquired from non-visual datasets.

Surprisingly, comparing the lists of these categories in case of the two detectors reveals that the same categories depends on the context the most (e.g.: snowboard, toothbrush, knife, baseball bet etc.). This suggests that these statistics are independent of the used detection technique and specific to the dataset, hence worth to be transferred.

Interestingly, there are case (e.g.: last row in Table~\ref{label_table_yolo_affected}) when masking of context improves the accuracy. This suggest a case when the detector was unsure whether the patch belongs to a \textit{notebook} or it is a separate object called \textit{keyboard}. Probably this is due to inconsistencies in the ground truth selecting procedure of the dataset.

\begin{table}
\centering
\caption{For categories with the largest contextual dependence, the accuracy drop of Faster-RCNN is reported. For each category, only those masked datasets are reported which belongs to the 3 largest deviations in accuracy. Datasets are labeled by the category which was masked out. Rows are sorted based on the largest accuracy deviation of the non self-masking case. Highlighted (yellow) cells indicate where masking context category affects the accuracy more than masking the object itself.}
\label{label_table_faster_affected}
\begin{tabular}{|
>{\columncolor[HTML]{C0C0C0}}l |l|c|c|c|}
\hline
\cellcolor[HTML]{EFEFEF}{\color[HTML]{333333} } & \cellcolor[HTML]{EFEFEF}{\color[HTML]{333333} } & \multicolumn{3}{c|}{\cellcolor[HTML]{EFEFEF}{\color[HTML]{333333} \textbf{\begin{tabular}[c]{@{}c@{}}Top-3 masked data sets for each \\ category which altered AP \\ the most (AP change in \%)\end{tabular}}}} \\ \cline{3-5} 
\multirow{-3}{*}{\cellcolor[HTML]{EFEFEF}{\color[HTML]{333333} \textbf{\begin{tabular}[c]{@{}l@{}}Categories\\ depending\\  on context\end{tabular}}}} & \multirow{-3}{*}{\cellcolor[HTML]{EFEFEF}{\color[HTML]{333333} \textbf{\begin{tabular}[c]{@{}l@{}}AP\\ for each\\ category\end{tabular}}}} & \cellcolor[HTML]{EFEFEF}{\color[HTML]{333333} 1st} & \cellcolor[HTML]{EFEFEF}{\color[HTML]{333333} 2nd} & \cellcolor[HTML]{EFEFEF}{\color[HTML]{333333} 3rd} \\ \hline
\textbf{snowboard} & 0,305 & \cellcolor[HTML]{FFFC9E}\begin{tabular}[c]{@{}c@{}}person\\ (34.8\%)\end{tabular} & \begin{tabular}[c]{@{}c@{}}skis\\ (84.3\%)\end{tabular} & \begin{tabular}[c]{@{}c@{}}skateboard\\ (91.4\%)\end{tabular} \\ \hline
\textbf{toothbrush} & 0,149 & \cellcolor[HTML]{FFFC9E}\begin{tabular}[c]{@{}c@{}}person\\ (48.6\%)\end{tabular} & \begin{tabular}[c]{@{}c@{}}toothbrush\\ (59.8\%)\end{tabular} & \begin{tabular}[c]{@{}c@{}}bottle\\ (106.3\%)\end{tabular} \\ \hline
\textbf{kite} & 0,356 & \begin{tabular}[c]{@{}c@{}}kite\\ (59.1\%)\end{tabular} & \begin{tabular}[c]{@{}c@{}}person\\ (63.4\%)\end{tabular} & \begin{tabular}[c]{@{}c@{}}dining\\ table (94.4\%)\end{tabular} \\ \hline
\textbf{cell phone} & 0,286 & \begin{tabular}[c]{@{}c@{}}cell \\ phone \\ (42.8\%)\end{tabular} & \begin{tabular}[c]{@{}c@{}}person\\ (65.2\%)\end{tabular} & \begin{tabular}[c]{@{}c@{}}cup\\ (97.7\%)\end{tabular} \\ \hline
\textbf{remote} & 0,231 & \begin{tabular}[c]{@{}c@{}}remote\\ (42.2\%)\end{tabular} & \begin{tabular}[c]{@{}c@{}}person\\ (66.5\%)\end{tabular} & \begin{tabular}[c]{@{}c@{}}tv\\ (97.6\%)\end{tabular} \\ \hline
\textbf{skis*} & 0,200 & \cellcolor[HTML]{FFFC9E}\begin{tabular}[c]{@{}c@{}}person\\ (70.9\%)\end{tabular} & \begin{tabular}[c]{@{}c@{}}skis\\ (75.5\%)\end{tabular} & \begin{tabular}[c]{@{}c@{}}backpack\\ (98.2\%)\end{tabular} \\ \hline
\textbf{handbag} & 0,110 & \begin{tabular}[c]{@{}c@{}}handbag\\ (32.2\%)\end{tabular} & \begin{tabular}[c]{@{}c@{}}person\\ (70.6\%)\end{tabular} & \begin{tabular}[c]{@{}c@{}}backpack\\ (103.1\%)\end{tabular} \\ \hline
\textbf{\begin{tabular}[c]{@{}l@{}}baseball\\ bat\end{tabular}} & 0,242 & \begin{tabular}[c]{@{}c@{}}baseball\\ bat\\ (53.0\%)\end{tabular} & \begin{tabular}[c]{@{}c@{}}person\\ (75.6\%)\end{tabular} & \begin{tabular}[c]{@{}c@{}}baseball\\ glove\\ (96.1\%)\end{tabular} \\ \hline
\textbf{knife*} & 0,122 & \cellcolor[HTML]{FFFC9E}\begin{tabular}[c]{@{}c@{}}dining\\ table\\ (76.6\%)\end{tabular} & \begin{tabular}[c]{@{}c@{}}knife\\ (80.6\%)\end{tabular} & \begin{tabular}[c]{@{}c@{}}person\\ (92.7\%)\end{tabular} \\ \hline
\textbf{surfboard} & 0,338 & \begin{tabular}[c]{@{}c@{}}surfboard\\ (75.4\%)\end{tabular} & \begin{tabular}[c]{@{}c@{}}person\\ (82.4\%)\end{tabular} & \begin{tabular}[c]{@{}c@{}}horse\\ (99.3\%)\end{tabular} \\ \hline
\textbf{frisbee} & 0,601 & \begin{tabular}[c]{@{}c@{}}frisbee\\ (40.8\%)\end{tabular} & \begin{tabular}[c]{@{}c@{}}person\\ (84.4\%)\end{tabular} & \begin{tabular}[c]{@{}c@{}}dog\\ (96.1\%)\end{tabular} \\ \hline
\textbf{fork} & 0,292 & \begin{tabular}[c]{@{}c@{}}fork\\ (75.7\%)\end{tabular} & \begin{tabular}[c]{@{}c@{}}dining table\\ (85.5\%)\end{tabular} & \begin{tabular}[c]{@{}c@{}}bowl\\ (97.8\%)\end{tabular} \\ \hline
\textbf{skateboard} & 0,478 & \begin{tabular}[c]{@{}c@{}}skateboard\\ (22.8\%)\end{tabular} & \begin{tabular}[c]{@{}c@{}}person\\ (86.6\%)\end{tabular} & \begin{tabular}[c]{@{}c@{}}bicycle\\ (98.8\%)\end{tabular} \\ \hline
\textbf{motorcycle} & 0,416 & \begin{tabular}[c]{@{}c@{}}motorcycle\\ (1.1\%)\end{tabular} & \begin{tabular}[c]{@{}c@{}}person\\ (86.2\%)\end{tabular} & \begin{tabular}[c]{@{}c@{}}bus\\ (99.2\%)\end{tabular} \\ \hline
\textbf{umbrella} & 0,342 & \begin{tabular}[c]{@{}c@{}}umbrella\\ (44.7\%)\end{tabular} & \begin{tabular}[c]{@{}c@{}}person\\ (88.5\%)\end{tabular} & \begin{tabular}[c]{@{}c@{}}chair\\ (101.0\%)\end{tabular} \\ \hline
\end{tabular}
\end{table}

\begin{table}
\centering
\caption{Similar to Table~\ref{label_table_faster_affected} in case of YOLO.}
\label{label_table_yolo_affected}
\begin{tabular}{|
>{\columncolor[HTML]{C0C0C0}}l |l|c|c|c|}
\hline
\cellcolor[HTML]{EFEFEF}{\color[HTML]{333333} } & \cellcolor[HTML]{EFEFEF}{\color[HTML]{333333} } & \multicolumn{3}{c|}{\cellcolor[HTML]{EFEFEF}{\color[HTML]{333333} \textbf{\begin{tabular}[c]{@{}c@{}}Top-3 masked data sets for each \\ category which altered AP \\ the most (AP change in \%)\end{tabular}}}} \\ \cline{3-5} 
\multirow{-3}{*}{\cellcolor[HTML]{EFEFEF}{\color[HTML]{333333} \textbf{\begin{tabular}[c]{@{}l@{}}Categories\\ depending \\ on context\end{tabular}}}} & \multirow{-3}{*}{\cellcolor[HTML]{EFEFEF}{\color[HTML]{333333} \textbf{\begin{tabular}[c]{@{}l@{}}AP\\ for each\\ category\end{tabular}}}} & \cellcolor[HTML]{EFEFEF}{\color[HTML]{333333} 1st} & \cellcolor[HTML]{EFEFEF}{\color[HTML]{333333} 2nd} & \cellcolor[HTML]{EFEFEF}{\color[HTML]{333333} 3rd} \\ \hline
\textbf{toothbrush} & 0,060 & \cellcolor[HTML]{FFFC9E}\begin{tabular}[c]{@{}c@{}}person\\  (22.1\%)\end{tabular} & \begin{tabular}[c]{@{}c@{}}toothbrush\\ (58.0\%)\end{tabular} & \begin{tabular}[c]{@{}c@{}}dining\\ table\\ (81.3\%)\end{tabular} \\ \hline
\textbf{snowboard} & 0,164 & \cellcolor[HTML]{FFFC9E}\begin{tabular}[c]{@{}c@{}}person\\ (60.0\%)\end{tabular} & \begin{tabular}[c]{@{}c@{}}snowboard\\ (87.7\%)\end{tabular} & \begin{tabular}[c]{@{}c@{}}skis\\ (93.7\%)\end{tabular} \\ \hline
\textbf{knife*} & 0,044 & \cellcolor[HTML]{FFFC9E}\begin{tabular}[c]{@{}c@{}}dining\\ table\\ (62.7\%)\end{tabular} & \begin{tabular}[c]{@{}c@{}}bowl\\ (105.2\%)\end{tabular} & \begin{tabular}[c]{@{}c@{}}cake\\ (95.4\%)\end{tabular} \\ \hline
\textbf{orange} & 0,049 & \begin{tabular}[c]{@{}c@{}}orange\\ (34.4\%)\end{tabular} & \begin{tabular}[c]{@{}c@{}}bowl\\ (65.8\%)\end{tabular} & \begin{tabular}[c]{@{}c@{}}apple\\ (86.1\%)\end{tabular} \\ \hline
\textbf{\begin{tabular}[c]{@{}l@{}}baseball\\ bat\end{tabular}} & 0,165 & \cellcolor[HTML]{FFFC9E}\begin{tabular}[c]{@{}c@{}}person\\ (70.2\%)\end{tabular} & \begin{tabular}[c]{@{}c@{}}baseball\\ bat\\ (82.9\%)\end{tabular} & \begin{tabular}[c]{@{}c@{}}chair\\ (96.6\%)\end{tabular} \\ \hline
\textbf{spoon} & 0,061 & \begin{tabular}[c]{@{}c@{}}spoon\\ (55.1\%)\end{tabular} & \begin{tabular}[c]{@{}c@{}}bowl\\ (70.9\%)\end{tabular} & \begin{tabular}[c]{@{}c@{}}dining\\ table\\ (108.1\%)\end{tabular} \\ \hline
\textbf{pizza} & 0,046 & \begin{tabular}[c]{@{}c@{}}pizza\\ (37.9\%)\end{tabular} & \begin{tabular}[c]{@{}c@{}}person\\ (70.9\%)\end{tabular} & \begin{tabular}[c]{@{}c@{}}dining\\ table\\ (115.7\%)\end{tabular} \\ \hline
\textbf{frisbee} & 0,349 & \begin{tabular}[c]{@{}c@{}}frisbee\\ (66.4\%)\end{tabular} & \begin{tabular}[c]{@{}c@{}}person\\ (74.1\%)\end{tabular} & \begin{tabular}[c]{@{}c@{}}dog\\ (95.7\%)\end{tabular} \\ \hline
\textbf{cell phone} & 0,181 & \begin{tabular}[c]{@{}c@{}}cell\\ phone\\  (37.2\%)\end{tabular} & \begin{tabular}[c]{@{}c@{}}person\\ (78.0\%)\end{tabular} & \begin{tabular}[c]{@{}c@{}}couch\\ (102.2\%)\end{tabular} \\ \hline
\textbf{\begin{tabular}[c]{@{}l@{}}baseball\\ glove\end{tabular}} & 0,137 & \begin{tabular}[c]{@{}c@{}}baseball\\ glove\\ (67.2\%)\end{tabular} & \begin{tabular}[c]{@{}c@{}}person\\ (79.5\%)\end{tabular} & \begin{tabular}[c]{@{}c@{}}baseball\\ bat\\ (104.6\%)\end{tabular} \\ \hline
\textbf{skateboard} & 0,325 & \begin{tabular}[c]{@{}c@{}}skateboard\\ (44.4\%)\end{tabular} & \begin{tabular}[c]{@{}c@{}}person\\ (80.5\%)\end{tabular} & \begin{tabular}[c]{@{}c@{}}traffic\\ light\\ (98.9\%)\end{tabular} \\ \hline
\textbf{hair drier} & 0,050 & \cellcolor[HTML]{FFFC9E}\begin{tabular}[c]{@{}c@{}}tv\\ (80.0\%)\end{tabular} & \begin{tabular}[c]{@{}c@{}}sink\\ (80.0\%)\end{tabular} & \begin{tabular}[c]{@{}c@{}}bottle\\ (80.0\%)\end{tabular} \\ \hline
\textbf{\begin{tabular}[c]{@{}l@{}}tennis\\ racket\end{tabular}} & 0,280 & \begin{tabular}[c]{@{}c@{}}tennis\\ racket\\ (35.6\%)\end{tabular} & \begin{tabular}[c]{@{}c@{}}person\\ (84.3\%)\end{tabular} & \begin{tabular}[c]{@{}c@{}}sports\\ ball\\ (97.3\%)\end{tabular} \\ \hline
\textbf{skis*} & 0,086 & \cellcolor[HTML]{FFFC9E}\begin{tabular}[c]{@{}c@{}}person\\ (84.9\%)\end{tabular} & \begin{tabular}[c]{@{}c@{}}skis\\ (86.5\%)\end{tabular} & \begin{tabular}[c]{@{}c@{}}backpack\\ (99.1\%)\end{tabular} \\ \hline
\textbf{remote} & 0,093 & \begin{tabular}[c]{@{}c@{}}remote\\ (52.2\%)\end{tabular} & \begin{tabular}[c]{@{}c@{}}person\\ (86.5\%)\end{tabular} & \begin{tabular}[c]{@{}c@{}}dining\\ table\\ (103.4\%)\end{tabular} \\ \hline
\textbf{keyboard} & 0,338 & \begin{tabular}[c]{@{}c@{}}keyboard\\ (18.4\%)\end{tabular} & \begin{tabular}[c]{@{}c@{}}laptop\\ (109.4\%)\end{tabular} & \begin{tabular}[c]{@{}c@{}}tv\\ (104.3\%)\end{tabular} \\ \hline
\end{tabular}
\end{table}

\section{Conclusion}\label{conclusion}

In MS COCO dataset there are object categories with different complexity.

State-of-the-art detectors can detect some object categories with high accuracy without relying on contextual information encoded at the level of co-occurring object categories.

However, there are some categories with ambivalent pixel patches which cannot be efficiently classified without looking at contextual information.
Measurements showed that contextual information described at the co-occurring object category level can hold relevant information to the classification.

Comparing the results of the two architectures revealed that despite the architectural differences the same object categories have the largest contextual dependence. This suggests that contextual information is independent of detection techniques and worth to be transferred.

\bibliographystyle{IEEEtran}
\bibliography{IEEEabrv,sample}

\end{document}